\documentclass[letterpaper, 10 pt, conference]{ieeeconf}  

\IEEEoverridecommandlockouts                              

\usepackage{graphics} 
\usepackage{graphicx}
\usepackage{amsmath} 
\usepackage{amssymb}  
\usepackage{booktabs} 
\usepackage{multirow}
\usepackage{caption}
\usepackage{xcolor}
\makeatletter
\let\NAT@parse\undefined
\makeatother
\usepackage{hyperref}

\title{\LARGE \bf
TASC: Task-Aware Shared Control for Relational Telemanipulation
}

\author{
Ze Fu$^{1,3}$, Pinhao Song$^{1,3}$, Yutong Hu$^{1,3}$, Renaud Detry$^{1,2,3}$
\thanks{$^{1}$KU Leuven, Dept. Mechanical Engineering, Research unit Robotics, Automation and Mechatronics, B-3000 Leuven, Belgium. {\tt\small firstname.lastname@kuleuven.be}}%
\thanks{$^{2}$KU Leuven, Dept. Electrical Engineering, Research unit Processing Speech and Images, B-3000 Leuven, Belgium.}
\thanks{$^{3}$Flanders Make@KU Leuven}
}

\begin{document}

\maketitle
\thispagestyle{empty}
\pagestyle{empty}

\begin{abstract}
We present TASC, a \underline{T}ask-\underline{A}ware \underline{S}hared \underline{C}ontrol framework for relational telemanipulation that infers task-level user intent and provides assistance from motion-only input. To support prehensile relational tasks without predefined templates, TASC constructs an open-vocabulary interaction graph from visual input to represent functional object relationships, and infers user intent accordingly. A shared control policy then provides assistance during both grasping and object interaction, guided by spatial constraints predicted by a vision-language model. Our method addresses two key challenges in relational telemanipulation under shared control: (1) task-level intent inference from low-level motion commands, and (2) generalizable assistance across diverse objects and tasks. Experiments in both simulation and the real world demonstrate that TASC improves task efficiency and reduces user input effort compared to prior methods, while enabling zero-shot generalization across diverse relational telemanipulation tasks. The code that supports our experiments is publicly available at \href{https://github.com/fitz0401/tasc}{\textcolor{blue}{https://github.com/fitz0401/tasc}}.
\end{abstract}

\section{INTRODUCTION}

Teleoperation remains essential in unstructured or safety-critical environments where full autonomy is unreliable due to perceptual uncertainty (e.g., underwater exploration~\cite{lin1997virtual}), task ambiguity (e.g., assistive robotics~\cite{liu2018motor}), or the need for human judgment (e.g., surgical robotics~\cite{ballantyne2003vinci}). However, directly controlling high-degree-of-freedom (DoF) manipulators is cognitively demanding and physically fatiguing, particularly for novice operators. Shared control mitigates this burden by blending human commands with autonomous assistance, improving precision and reducing cognitive load~\cite{xu2022SA_BCI, Gillini2021SA_BCI}. Despite promising results, existing shared control methods primarily provide simple goal-reaching~\cite{Song2024RobotTrajectron, Shervin2018HO, Gopinath2017HumanintheLoop} and single-step grasping~\cite{Xu2022graspandreach, Yow2024SAPOMDP} assistance. Extending these capabilities to task-aware assistance across diverse everyday manipulation scenarios remains challenging.

A central challenge in shared control is inferring user intent from available input signals. Prior approaches incorporate multimodal cues such as language for high-level goal specification~\cite{Yow2024SAPOMDP}, eye gaze for object attention~\cite{admoni2016predicting}, and motion commands from teleoperation devices~\cite{Song2024RobotTrajectron}. In this work, we focus on intent conveyed solely through motion input. This design choice reflects a practical constraint in many real-world teleoperation settings, where motion commands constitute the most accessible and widely available communication channel. 

We specifically consider prehensile relational tasks, in which manipulation unfolds as a sequence of object grasping followed by functional object-object interaction (e.g., insertion, placing, pouring). Such tasks require reasoning not only about spatial motion but also about task progression and inter-object constraints. Current motion-based shared control methods struggle in this setting. We identify two key challenges:

\noindent\textbf{(1) Task-level intent inference.} Inferring user intent from motion commands is difficult in relational manipulation tasks. For example, pouring water involves grasping a kettle, moving it over a cup, and executing the pour. To support such tasks, shared control systems must identify the current task and recognize task progression.
\noindent\textbf{(2) Generalizable assistance across tasks and objects.} A practical shared control system should operate across diverse tasks and objects without predefined task templates. Existing approaches are typically tailored to well-structured scenarios, limiting their applicability to everyday manipulation tasks.

\begin{figure}[tb]
    \centering
    \includegraphics[width=0.95\linewidth]{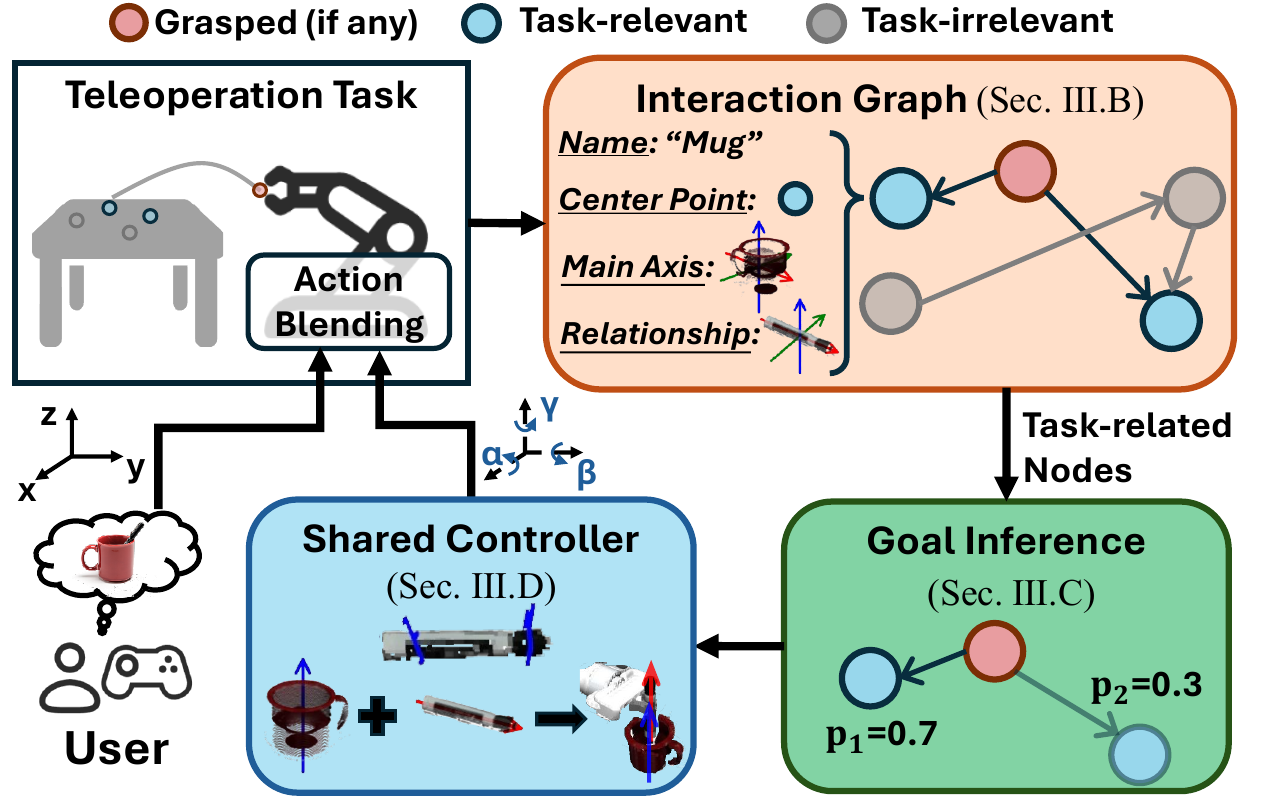}
    \caption{\rmfamily Overview of our TASC framework. The teleoperation loop begins with RGB-D observations of the scene, which are processed to construct an interaction graph (see Sec.~\ref{sec:graph}) encoding object centers, main axes, and functional relationships. In the interaction graph, the \textcolor{red}{red} node denotes the grasped object (if any), \textcolor{blue}{blue} nodes denote task-relevant objects that serve either as grasp targets or as functionally related objects, and \textcolor{gray}{gray} nodes denote task-irrelevant objects. The inference module (see Sec.~\ref{sec:prediction}) fuses this graph with motion cues to estimate a goal distribution. Guided by this distribution, the shared controller (see Sec.~\ref{sec:policy}) blends the user's translational input with assistive rotations.}
    \label{fig:framework}
\end{figure}

To tackle these challenges, shared control systems require flexible representations of functional object relationships to modulate assistance according to task progression, rather than rely on predefined task templates and assistance primitives. Recent advances in vision-language models (VLMs) demonstrate the feasibility of extracting semantic and functional relations from visual scenes~\cite{huang2024rekep, pan2025omnimanip, jiang2024roboexp, yan2025dynamic}. Inspired by these advances, we propose TASC (see Fig.~\ref{fig:framework}), a task-aware shared control framework that supports task-level intent inference and assistance throughout task execution. Our system consists of two key components:
\\
(1) First, we construct an \textbf{open-vocabulary interaction graph} capturing possible object-object relationships and corresponding interaction constraints. This graph representation helps our system to understand the composition and progression of multi-step prehensile relational tasks. 
\\
(2) Second, we present a \textbf{multi-step shared control policy} that unifies grasping and subsequent object interaction within a single optimization framework. During grasping, the system assists the user by selecting an optimal grasp pose. During subsequent object interaction, it infers spatial affordances between the grasped object and its target via the interaction graph, and formulates an optimization problem to solve for a suitable pose for interaction.

Our experiments, with both simulated and real-world teleoperated manipulation, show that TASC allows users to complete multi-step tasks with reduced input burden and improved efficiency, and generalizes across multiple tasks and object configurations without task-specific training.

The main contributions of this paper are:
\begin{itemize}
    \item A task-aware shared control framework that represents functional object relationships through an open-vocabulary interaction graph, enabling task-level user intent inference from motion-only input without predefined task templates.
    \item A shared control policy that unifies grasping and subsequent object interaction, modulating assistance according to task progression and selectively adjusting task-relevant degrees of freedom.
    \item Comprehensive evaluations in simulation and real-world teleoperation, demonstrating improved efficiency, reduced user input, and generalization across diverse multi-step manipulation tasks compared to pure teleoperation and shared control baselines.
\end{itemize}

\section{RELATED WORK}

\textbf{Shared Control} for high-DoF robotic manipulators has been widely explored to reduce user burden during teleoperation. Early approaches primarily provide end-effector (EEF) translation assistance. For instance, planning-based methods employ artificial potential fields to steer the robot toward likely targets~\cite{Gottardi2022APF}, while learning-based approaches predict future trajectories from user motion history to enable proactive support~\cite{Song2024RobotTrajectron, yoneda2023noise}. To provide rotation assistance, which is often the most unintuitive aspect of teleoperation, prior works typically rely on predefined goal objects~\cite{Shervin2018HO, Yow2024SAPOMDP, dragan2013policy, nikolaidis2017human} or discrete action templates~\cite{Gabriel2020templates}. Although effective in structured settings, these methods generally limited to short-horizon behaviors such as reaching or grasping~\cite{Xu2022graspandreach, Gopinath2017HumanintheLoop}. More recent efforts begin incorporating scene context for task-level assistance. For example, VOSA~\cite{Atharv2025vosa} leverages vision-based object detection to support zero-shot intent recognition. However, existing approaches provide limited support for post-grasp object interactions. In contrast, our work introduces an interaction-aware representation to enable task-level assistance that adapts across different multi-step manipulation tasks.

\textbf{Graph-based Representations} are well-suited for modeling object relationships. In indoor navigation, scene graphs have been used to represent spatial layouts and semantic relationships to support language-guided path planning~\cite{yan2025dynamic}. In manipulation, action-conditioned scene graphs~\cite{jiang2024roboexp} have enabled interactive exploration and planning over long-horizon tasks by capturing how object states and relationships evolve in response to robot actions. Compared to prior works~\cite{yan2025dynamic, jiang2024roboexp, shang2025sg, zhu2021hierarchical} that primarily model spatial relations (e.g., on, inside), we focus on capturing potential functional interactions between object pairs. In this direction, prior efforts~\cite{unmesh2023interacting, mo2022o2o} in object-object interactions (OOI) have proposed taxonomies and datasets to classify spatio-temporal relationships between objects. However, their applications in robotic manipulation remain underexplored.

\textbf{Affordance for Manipulation} provides actionable priors that support tool usage and motion planning. Early works~\cite{chu2019learning, fang2020learning, xu2021affordance} focus on learning single-object affordances through data collection and supervised training, while recent efforts~\cite{huang2024imagination, qi2025two} extend to modeling object-object affordances. However, these approaches are constrained by data collection efficiency and limited generalization. The emergence of foundation models has made zero-shot affordance inference increasingly feasible because of their ability for semantic scene understanding and commonsense reasoning. Recent works~\cite{liu2024moka, huang2024copa, huang2024rekep} use VLMs to extract keypoint constraints between objects under task instructions. However, due to the limited spatial grounding capabilities of current VLMs, the resulting keypoints can be unstable or inconsistent across queries. Inspired by OmniManip~\cite{pan2025omnimanip}, we instead extract task-relevant pose alignment constraints between object pairs via VLMs. This provides a stable and concise representation that is well-suited for assisting rotation in shared control.

\section{METHOD}

We address multi-step prehensile relational teleoperation under shared control, where a user operates a high-DoF robotic manipulator through low-dimensional motion input. To this end, we introduce TASC (see Fig.~\ref{fig:framework}), a framework consisting of: (1)~an open-vocabulary interaction graph that encodes functional object relationships (Sec.~\ref{sec:graph}); (2)~a goal inference module that maintains a belief over the user's task-level intent (Sec.~\ref{sec:prediction}); and (3)~a shared control policy that provides rotation assistance during both grasping and post-grasp object interaction (Sec.~\ref{sec:policy}).

\subsection{Problem Formulation}
We consider a scene containing $N$ objects and a finite but unspecified task space $\mathcal{T}$, where each task corresponds to a functionally meaningful object-object interaction (e.g., placing a banana on a plate). The user intends to complete one task in $\mathcal{T}$, which is unknown to the system. We denote the robot state by $x$, consisting of the EEF position and orientation. During teleoperation, the user provides EEF motion commands, including translational input $u_\text{h}^t \in \mathbb{R}^3$ and optional rotational input $u_\text{h}^r \in \mathbb{R}^3$. The system generates assistive rotational commands $u_\text{a}^{\mathrm{r}} \in \mathbb{R}^3$ to support task execution. The final command executed by the robot is $u = [(u_\text{h}^{\mathrm{t}})^\top, (u_\text{h}^{\mathrm{r}} + u_\text{a}^{\mathrm{r}})^\top ]^\top $. Our objective is to infer the user's task-level intent from motion input, and to generate appropriate rotational commands in a zero-shot manner.

\subsection{Open-vocabulary Interaction Graph}
\label{sec:graph}
\begin{figure}[tb]
    \centering
    \includegraphics[width=0.95\linewidth]{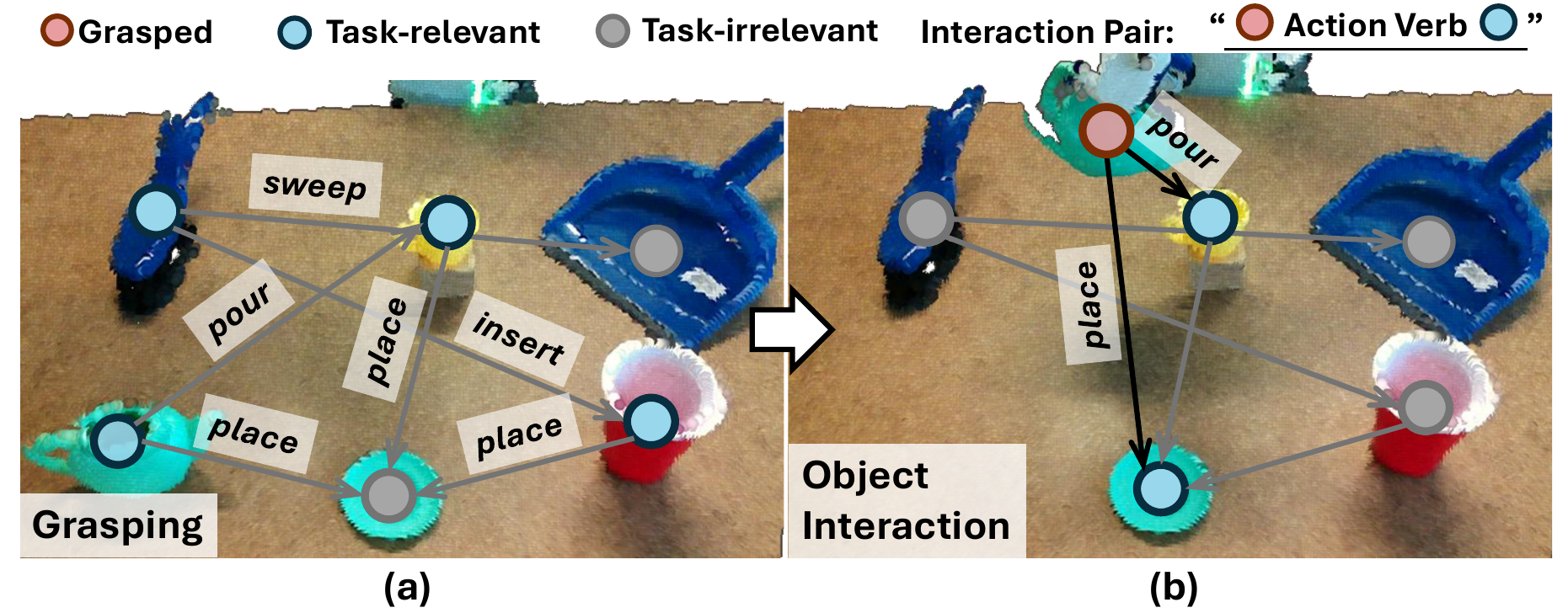}
    \caption{\rmfamily The interaction graph with nodes in different activation states: (a) During grasping, nodes with outgoing edges (i.e., functional tools) are marked active; (b) during object interaction, only the neighbors of the grasped node are considered.}
    \label{fig:scene_graph}
\end{figure}

To enable task-level reasoning without predefined task templates, we construct an \emph{open-vocabulary interaction graph} that encodes functional relationships between objects (Fig.~\ref{fig:scene_graph}). An interaction is defined as a functionally meaningful triplet of the form \emph{object\_A – action – object\_B} (e.g., ``hammer–hit–nail'').

Given an RGB observation $\mathcal{R}$ containing objects $\mathcal{O}=\{o_i\}_{i=1}^{N}$, we query a VLM to extract functionally relevant object names and candidate interaction triplets. Rather than exhaustively detecting all objects, we restrict extraction to objects that participate in potential interactions, avoiding the need to tune recognition granularity manually. We then construct a directed graph $\mathcal{G}=(V,E)$, where each node $v_i \in V$ corresponds to an object $o_i$, and each edge $e_{ij} \in E$ represents a candidate interaction from $o_i$ to $o_j$ associated with an action verb.

To ground semantic relations geometrically, we use GroundedSAM~\cite{ren2024grounded} to obtain object masks and extract spatial information from RGB-D input. For each object, we compute its 3D position and point cloud, which are stored in the corresponding node. The resulting interaction graph serves as a compact representation of $\mathcal{T}$, capturing both semantic affordances and geometric grounding for downstream intent inference and shared control.

\subsection{Goal Inference over Interaction Graph}
\label{sec:prediction}
To generalize across diverse prehensile relational tasks, we decompose multi-step manipulation tasks into two stages---\textbf{grasping} and \textbf{object interaction}---and infer the operation goal in each stage. Unlike prior methods that treat all visible objects as potential goals~\cite{Yow2024SAPOMDP, Atharv2025vosa}, we restrict the candidate goal set using the interaction graph to reduce ambiguity in cluttered scenes. During grasping, only objects with non-zero out-degree are considered, representing objects that can act on others. During object interaction, the candidate set $G$ contains only objects directly connected to the currently grasped object.

At time step $t$, given the sequence of robot states and user inputs $\xi^{0:t} = \{x^0, u_\text{h}^0, \dots, x^t, u_\text{h}^t\}$, we estimate the posterior probability over candidate goals:
\begin{equation}
\label{equ: bayes}
    p(g | \xi^{0:t}) \propto p(\xi^{0:t} | g)p(g)
\end{equation}

Assuming conditional independence of $u_\text{h}^t$ given $x^t$ and $g$, the posterior can be updated recursively:
\begin{equation}
\label{equ: p(g|xi)}
    p(g | \xi^{0:t}) \propto p(\xi^{0:t-1} | g)p(u_\text{h}^t | x^t, g)
\end{equation}

We model the likelihood $p(u_\text{h}^t \mid x^t, g)$ using a maximum entropy inverse optimal control (MaxEnt IOC) formulation~\cite{ziebart2008maximum, Shervin2018HO}, where user actions are assumed to minimize a cost defined by the Euclidean distance between the current EEF position and goal object $g$. Lower-cost actions are exponentially more likely, yielding a smooth probabilistic estimate of intent. The resulting belief distribution $p(g \mid \xi^{0:t})$ is updated online and guides the downstream shared control policy. Notably, we provide assistance only in rotational DoF to avoid feedback bias in translational intent inference.

\subsection{Shared Control Policy}
\label{sec:policy}

\textbf{Grasping Assistance.} During grasping, we assist orientation to reduce the difficulty of precise alignment. For the most probable object $g^* = \arg\max p(g \mid \xi^{0:t})$, we generate candidate grasp poses using a pretrained grasp planner~\cite{fang2023anygrasp}. To ensure both diversity and spatial coverage, we apply farthest point sampling to dense predictions. The grasp pose requiring minimal rotational adjustment from the current EEF pose is selected as the assistive target. Rotational commands are interpolated toward this target, ensuring lightweight and intuitive guidance.

\begin{figure}[tb]
    \centering
    \includegraphics[width=0.9\linewidth]{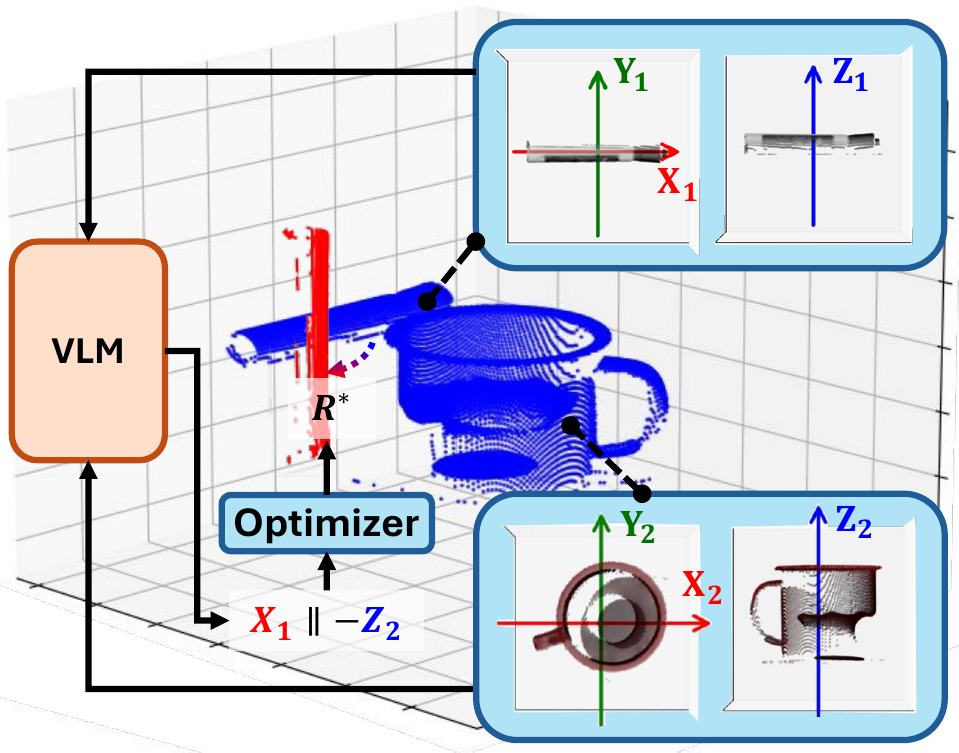}
    \caption{\rmfamily Object interaction assistance through functional axis alignment. Given the inferred interaction pair, the system predicts alignment constraints between object axes and computes the optimal rotation to satisfy them.}
    \label{fig:spatial_affordance}
\end{figure}

\textbf{Object Interaction Assistance.} After a successful grasp, assistance shifts to enforcing functional pose relationships between the grasped object $o_{\text{src}}$ and its inferred target $o_{\text{tgt}}$ (Fig.~\ref{fig:spatial_affordance}). From the interaction graph, we retrieve the predicted interaction pair and compute canonical axes for both objects using oriented bounding boxes. We then render axis-overlaid object images and prompt a VLM to infer axis alignment constraints between the two objects, yielding a set of constraints $\{(a_i, b_i, s_i) \}_{i=1}^k$, which specify that axis $a_i$ of $o_{\text{src}}$ should be aligned (if $s_i = 1$) or anti-aligned (if $s_i = -1$) with axis $b_i$ of $o_{\text{tgt}}$.  Given these constraints, we compute the optimal rotation $R^* \in SO(3)$ that minimizes axis misalignment:
\begin{equation}
    R^* = \arg\min_{R \in SO(3)} \sum_i \|Ra_i - s_i b_i\|^2.
\end{equation}
The resulting rotation is blended with user translation to support precise object-object alignment. This design enables task-aware assistance beyond grasping, supporting diverse post-grasp interactions without predefined task templates.

\section{EXPERIMENTS}
To comprehensively evaluate the performance of TASC, we conducted both simulation studies and real-world shared control experiments. All experiments were performed on the Franka Research 3 robotic platform. For our method, we used Qwen-VL~\cite{Qwen-VL} to extract task-relevant object relations for constructing the interaction graph. Prompts used for querying the model are provided in our code repository. All user studies were approved by the Social and Societal Ethics Committee of KU Leuven (G-2024-8370). The collected data were processed in accordance with the General Data Protection Regulation (GDPR) of the EU.

\subsection{Experiment in the Simulation}
We evaluate our method to answer two main questions: (1) Can TASC achieve accurate task-level intent inference? (2) Can TASC improve efficiency and reduce input effort?

\textbf{Simulation Setup.} We construct a tabletop teleoperation environment in Robosuite~\cite{robosuite2020} with six YCB~\cite{ycb2015} objects. Three multi-step tasks of increasing orientation complexity are defined: (1) \textit{\textbf{Banana-Plate}:} place a banana onto a plate. (2) \textit{\textbf{Marker-Mug}:} insert a marker into a mug. (3) \textit{\textbf{Hammer-Peg}:} strike a peg with a hammer. All tasks involve both grasping and post-grasp object interaction. The user and the system both observe the environment from a single calibrated top-down RGB-D view. Notably, our system has no access to any prior knowledge about object properties or task details.

\begin{figure}[tb]
    \centering
    \includegraphics[width=0.9\linewidth]{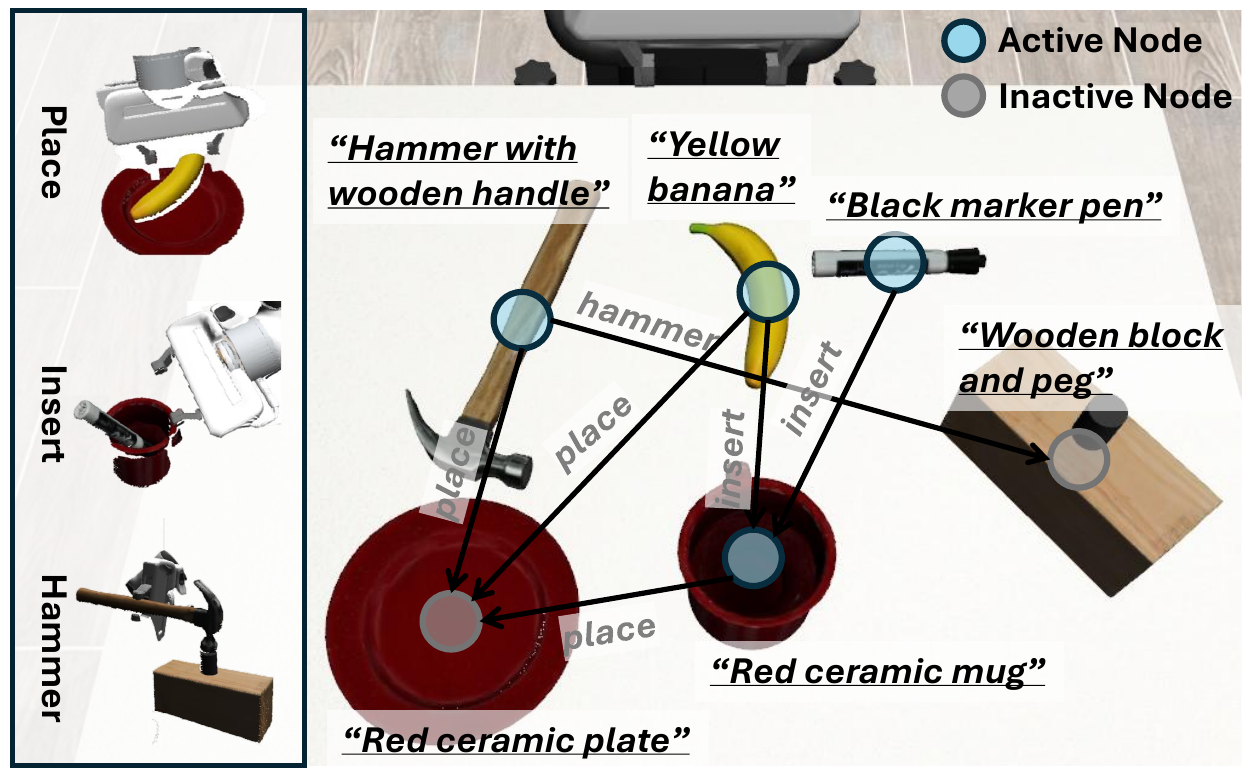}
    \caption{\rmfamily Simulation setup and representative initial object configurations for multi-step shared control tasks.}
    \label{fig:simulation_setting}
\end{figure}

\textbf{Protocol.} The environment provides a teleoperation interface enabling users to control the EEF position, orientation, and the gripper state via keyboard. Each trial starts with a fixed initial EEF pose, while object poses are randomized. We recruited 10 novice participants from the local community. Each received brief training to operate the robot using a keyboard. During the formal experiment, each participant completed all three tasks under each shared control method in randomized order.

\textbf{Baselines and Metrics.} 
We compared TASC with three baselines. 
\begin{itemize}
    \item \textbf{Teleop:} pure user control without assistance.
    \item \textbf{TASC$^{-}$:} an ablation of our method without using the interaction graph. All graspable objects are treated as candidate goals during grasping, and all remaining objects are considered potential targets during interaction.
    \item \textbf{VOSA~\cite{Atharv2025vosa}:} a vision-based shared control method that provides intent recognition and position assistance for pick-and-place tasks. To minimize comparison bias, we implement the same perception module (GroundedSAM~\cite{ren2024grounded}) for VOSA as in TASC.
\end{itemize}
For each trial, we recorded the completion time, trajectory length of the EEF, and the number of control inputs (button presses) during the task execution.

\begin{table}[t]
\centering
\caption{Performance across teleoperation tasks. Each cell shows (Time (s) / Trajectory Length (m) / Inputs).}
\label{tab:sim_results}
\setlength{\tabcolsep}{4pt}
\begin{tabular}{l|c|c|c}
    \toprule
    Method & Banana--Plate & Marker--Mug & Hammer--Peg \\
    \midrule
    Teleop
        & \textbf{54.7} / \textbf{1.50} / 135.4
        & 99.9 / 2.16 / 228.3
        & 112.9 / 3.40 / 338.6 \\
    VOSA 
        & 61.9 / 1.57 / 136.9
        & 97.8 / 2.41 / 222.0
        & 132.9 / 3.80 / 341.2 \\
    TASC$^{-}$
        & 91.0 / 2.26 / 178.7
        & 124.5 / 2.99 / 267.9
        & 73.9 / \textbf{2.42} / 218.0 \\
    TASC 
        & 66.4 / 1.71 / \textbf{116.6}
        & \textbf{54.7} / \textbf{1.52} / \textbf{95.6}
        &  \textbf{72.1} / \textbf{2.42} / \textbf{186.7}\\
    \bottomrule
\end{tabular}
\end{table}

\textbf{Results and Analysis.} 
Results are summarized in Table~\ref{tab:sim_results}. TASC consistently achieves fewer control inputs across all tasks, indicating lower user burden. In the simple \emph{banana--plate} task which primarily involves straightforward reaching, performance differences are modest. For the more challenging \emph{marker--mug} and \emph{hammer--peg} tasks involving precise alignment, TASC achieves shorter completion time and reduced EEF trajectory length, suggesting that task-aware rotational assistance is beneficial. Compared to TASC$^{-}$, the full model demonstrates improved efficiency, indicating the importance of interaction-graph-based goal filtering. Compared to VOSA, TASC achieves better performance in tasks involving post-grasp object interaction, highlighting the benefit of modeling object-object relationships.

\begin{figure}[tb]
    \centering
    \includegraphics[width=0.9\linewidth]{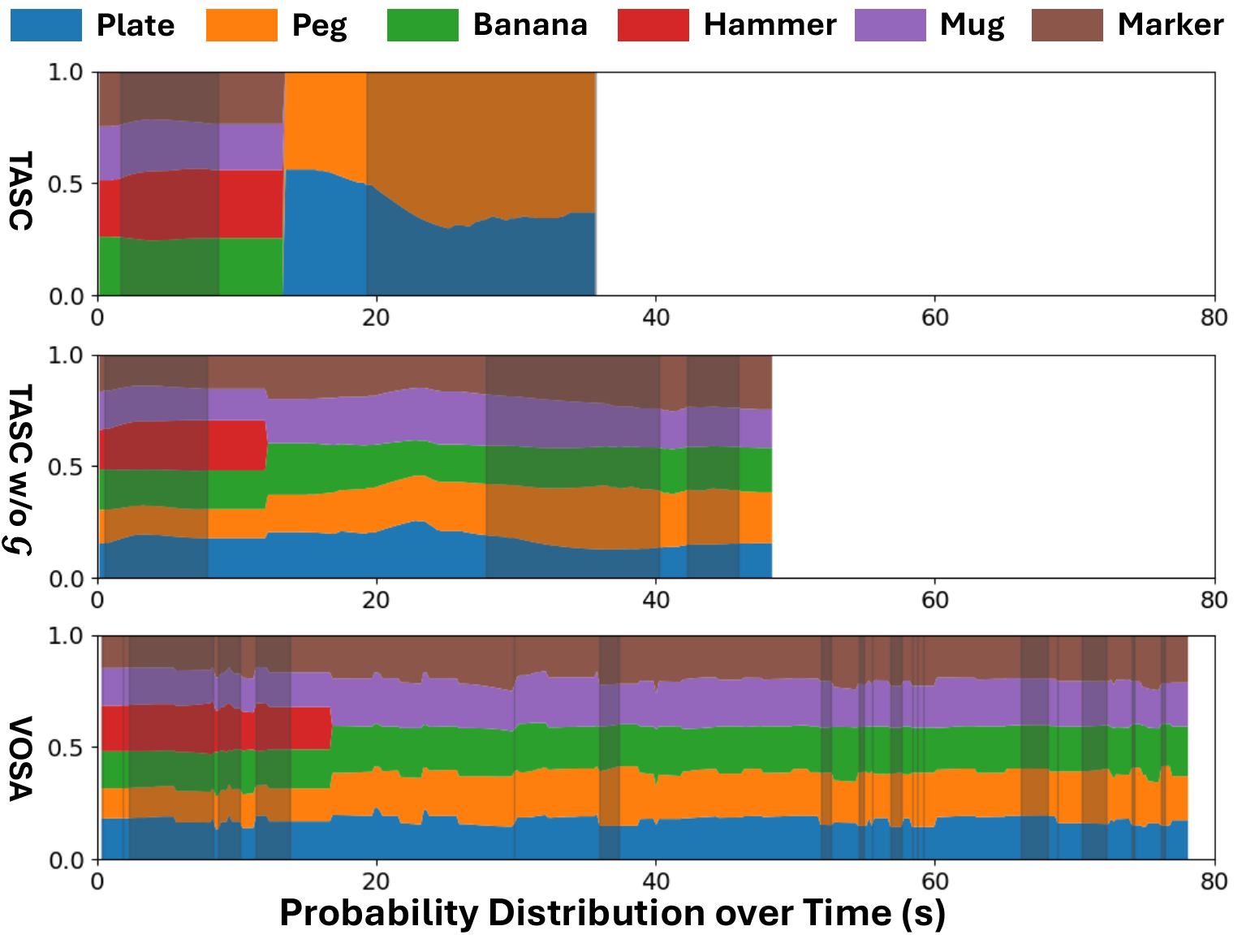}
    \caption{\rmfamily Goal probability distribution over time in \emph{hammer--peg} task across three shared control methods; shaded regions denote correct goal predictions during the grasping and object interaction phases.}
    \label{fig:prob_distribution}
\end{figure}

We further analyze intent inference by visualizing the evolution of $p(g \mid \xi^{0:t})$ in Fig.~\ref{fig:prob_distribution} using a representative \emph{hammer--peg} trial. Each colored area represents the probability assigned to a candidate goal, while shaded regions indicate periods in which the system correctly identifies the intended object. Our method exhibits early convergence to the correct goal and maintains stable predictions throughout. In contrast, TASC$^{-}$ and VOSA show delayed convergence and fluctuating predictions, particularly during the object interaction stage. This indicates that our intent inference mechanism improves both inference accuracy and temporal stability, which contributes to smoother assistance.

\subsection{Experiment in the Real-World}
We conduct real-world experiments to evaluate: (1) the robustness and generalization of TASC across diverse tasks, and (2) its effectiveness in improving user performance and experience during teleoperation.

\begin{figure}[tb]
    \centering
    \includegraphics[width=0.95\linewidth]{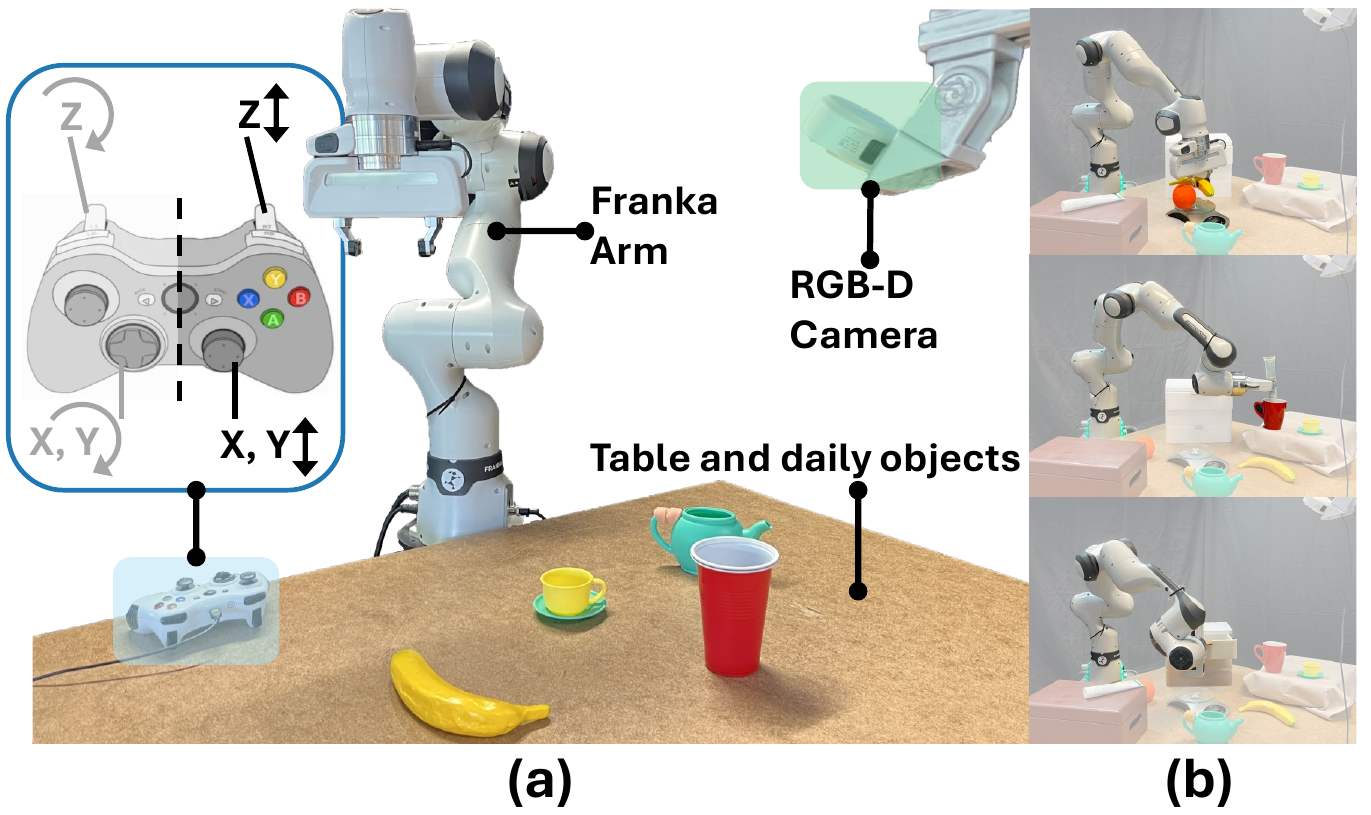}
    \caption{\rmfamily Real-world experimental setup. (a) Tabletop workspace with a Franka Research 3 arm and a top-down RGB-D camera. (b) Three representative tasks: Scale-Fruits, Insert-Tube, and Open-Drawer (from top to bottom).}
    \label{fig:real_world_setting}
\end{figure}

\textbf{Settings.} Experiments were performed in a tabletop workspace (Fig.~\ref{fig:real_world_setting}), with a RealSense L515 RGB-D camera providing a top-down view. We prepared diverse object configurations and interaction pairs, and designed three representative tasks: (1) \textit{\textbf{Scale-Fruits}:} place a banana and an orange onto a weighing scale. (2) \textit{\textbf{Insert-Tube}:} insert a tube into a mug. (3) \textit{\textbf{Open-Drawer}:} open a hinged drawer. 

\begin{figure}[tb]
    \centering
    \includegraphics[width=1.0\linewidth]{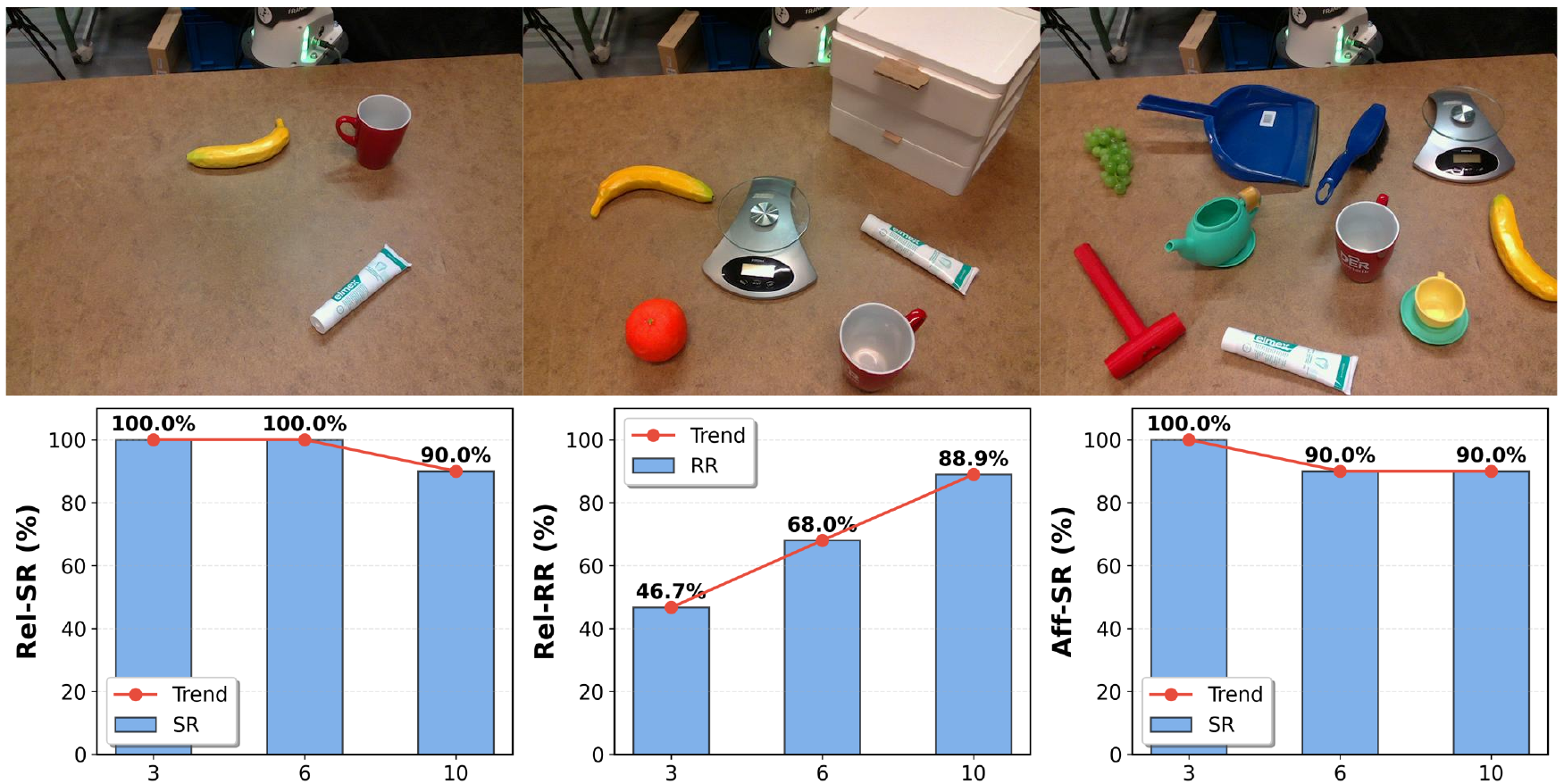}
    \caption{\rmfamily Robustness evaluation. Top row: Example scenes with increasing object counts (3, 6, and 10 objects), with 10 randomized trials per setting. Bottom row: Rel-SR, Rel-RR, and Aff-SR (from left to right) under varying object counts.}
    \label{fig:robustness}
\end{figure}

\textbf{Baselines and Metrics.} To evaluate robustness of interaction graph construction (Sec.~\ref{sec:graph}), we prepared scenes with 3, 6, and 10 everyday objects under randomized configurations (Fig.~\ref{fig:robustness}). A graph is considered correct only if it recovers both (1) the fruit-place–scale and tube-insert–mug interaction relationships and (2) the corresponding alignment constraints. Each setting is evaluated over 10 trials. We report relationship inference success rate (Rel-SR) and affordance inference success rate (Aff-SR). We also report the relationship reduction rate (Rel-RR), defined as the proportion of irrelevant object pairs pruned by interaction reasoning. Given $n$ objects and $k$ retained relations, Rel-RR is computed as $1 - \frac{k}{n(n-1)}$.

We further evaluate teleoperation performance using correctly inferred interaction graphs. We compare TASC with:
\begin{itemize}
    \item \textbf{Teleop}: pure user control without assistance.
    \item \textbf{MaxEnt IOC}~\cite{Shervin2018HO, Gottardi2022APF}: a shared control method that provides full pose assistance toward goals. For fair comparison, it uses the same grasp planner as TASC, while it does not model post-grasp object-object interaction constraints.
\end{itemize}
Ten novice participants were recruited and trained briefly to operate the robot via joystick. Each participant performed all tasks under all methods in randomized order. We report the task success rate and completion time. Note that a trial is considered failed if the robot enters an unrecoverable configuration (e.g., kinematic deadlock) or causes irreversible environment disturbance (e.g., mug toppled). Participants additionally rated five subjective measures (0–10 scale):
\begin{itemize}
    \item Control: I felt in control of the robot. 
    \item Expectation: The robot did what I wanted.
    \item Speed: I was able to complete the tasks quickly.
    \item Usability: I would like to use this system for future teleoperation tasks.
    \item Satisfaction: I was satisfied with my performance.
\end{itemize}

\textbf{Results.} Robustness results are shown in Fig.~\ref{fig:robustness}. Rel-SR remains stable as the number of objects increases, indicating reliable inference in cluttered scenes. Aff-SR also shows no degradation, benefiting from object segmentation. The failure cases are caused by severe occlusion leading to incomplete point clouds. We further observe that interaction analysis becomes more beneficial in clustered scenes. For example, with 10 objects, 88.9\% of potential object pairs are pruned while preserving the required insertion and placing relations.

Results of user performance are summarized in Table~\ref{tab:real_world_results}. TASC achieves high assistance success across all tasks and attains the fastest execution time in two of the three scenarios. For Scale–Fruits, MaxEnt IOC shows no speed improvement in clutter due to distraction from irrelevant objects. In contrast, TASC accelerates execution by retaining only graspable objects. For Insert–Tube, TASC substantially improves success rate and completion time through insertion assistance. For Open–Drawer, which does not involve multi-object interaction reasoning, MaxEnt IOC is faster due to full-pose assistance, while TASC achieves comparable performance. Overall, these results indicate that our method particularly benefits cluttered and alignment-critical tasks.

\begin{table}[t]
\centering
\caption{Teleoperation performance in three real-world tasks. Each cell reports success rate and execution time (s).}
\label{tab:real_world_results}
\setlength{\tabcolsep}{4pt}
\begin{tabular}{l|c|c|c}
    \toprule
    Method & Scale--Fruits & Insert--Tube & Open--Drawer \\
    \midrule
    Teleop
        & \textbf{0.9} / 84.8 ± 26.0
        & 0.6 / 78.3 ± 45.7
        & 0.8 / 71.3 ± 20.6\\
    MaxEnt IOC 
        & \textbf{0.9} / 87.0 ± 28.4
        & 0.6 / 75.6 ± 34.2
        & \textbf{1.0} / \textbf{35.3 ± 13.7} \\
    TASC 
        & \textbf{0.9} / \textbf{77.9 ± 20.2}
        & \textbf{0.9} / \textbf{65.5 ± 26.6}
        & \textbf{1.0} / 41.5 ± 15.8 \\
    \bottomrule
\end{tabular}
\end{table}

\begin{figure}[tb]
    \centering
    \includegraphics[width=1.0\linewidth]{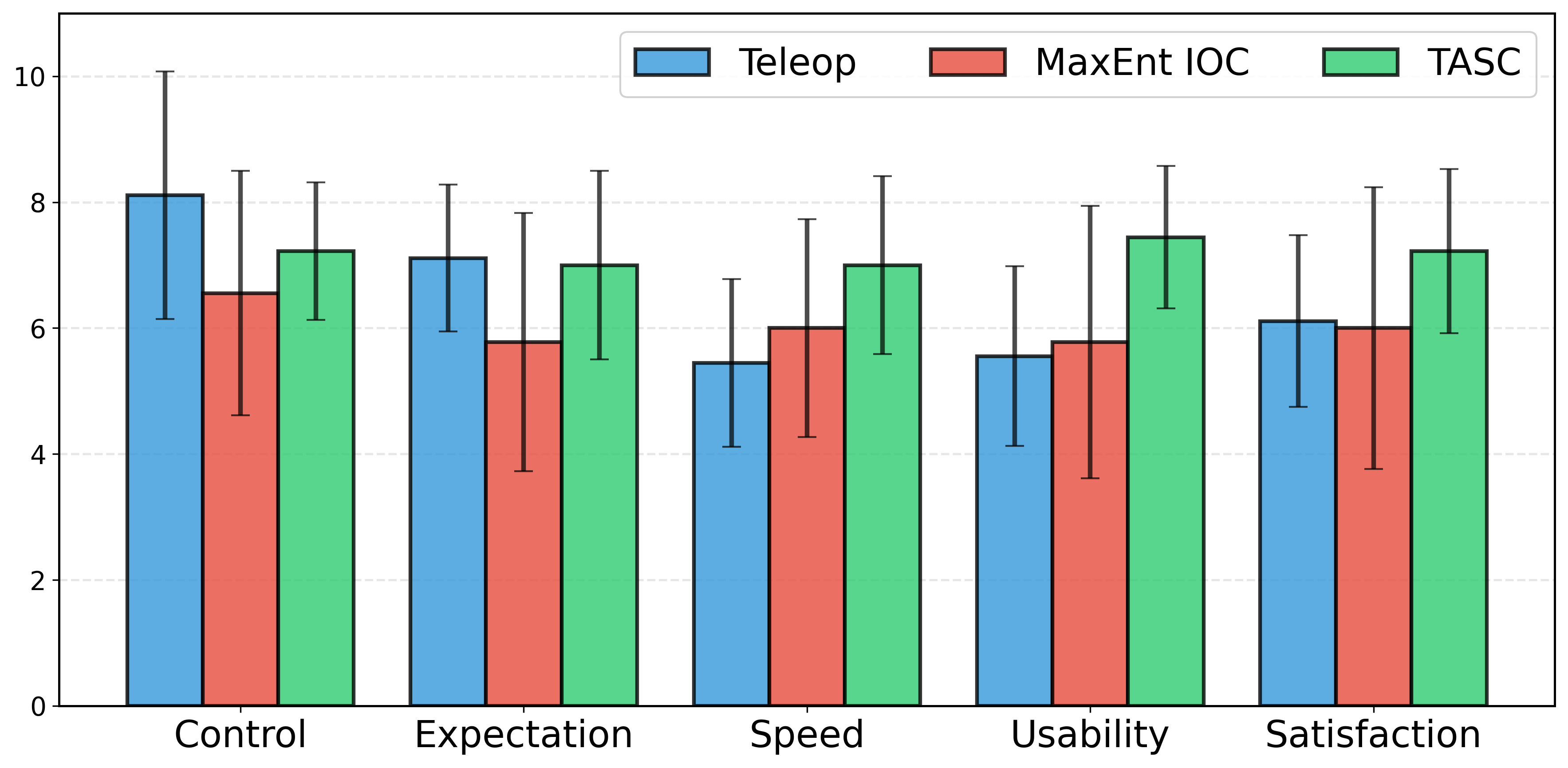}
    \caption{\rmfamily User study results across five subjective metrics: control, expectation alignment, speed, usability, and satisfaction.}
    \label{fig:subjective_metrics}
\end{figure}

As illustrated in Fig.~\ref{fig:subjective_metrics}, TASC achieves higher mean ratings on most subjective metrics compared to the baselines. While pure teleoperation naturally provides the highest sense of control, TASC maintains a comparable level of perceived control and how well the robot behaves as expected, demonstrating a favorable balance between user control and assistance. Paired t-tests reveal that TASC yields statistically significant improvements over pure teleoperation in Speed ($p=0.028$), Usability ($p=0.012$), and Satisfaction ($p=0.030$), confirming its effectiveness in enhancing the overall teleoperation experience.

\begin{figure}[tb]
    \centering
    \includegraphics[width=1.0\linewidth]{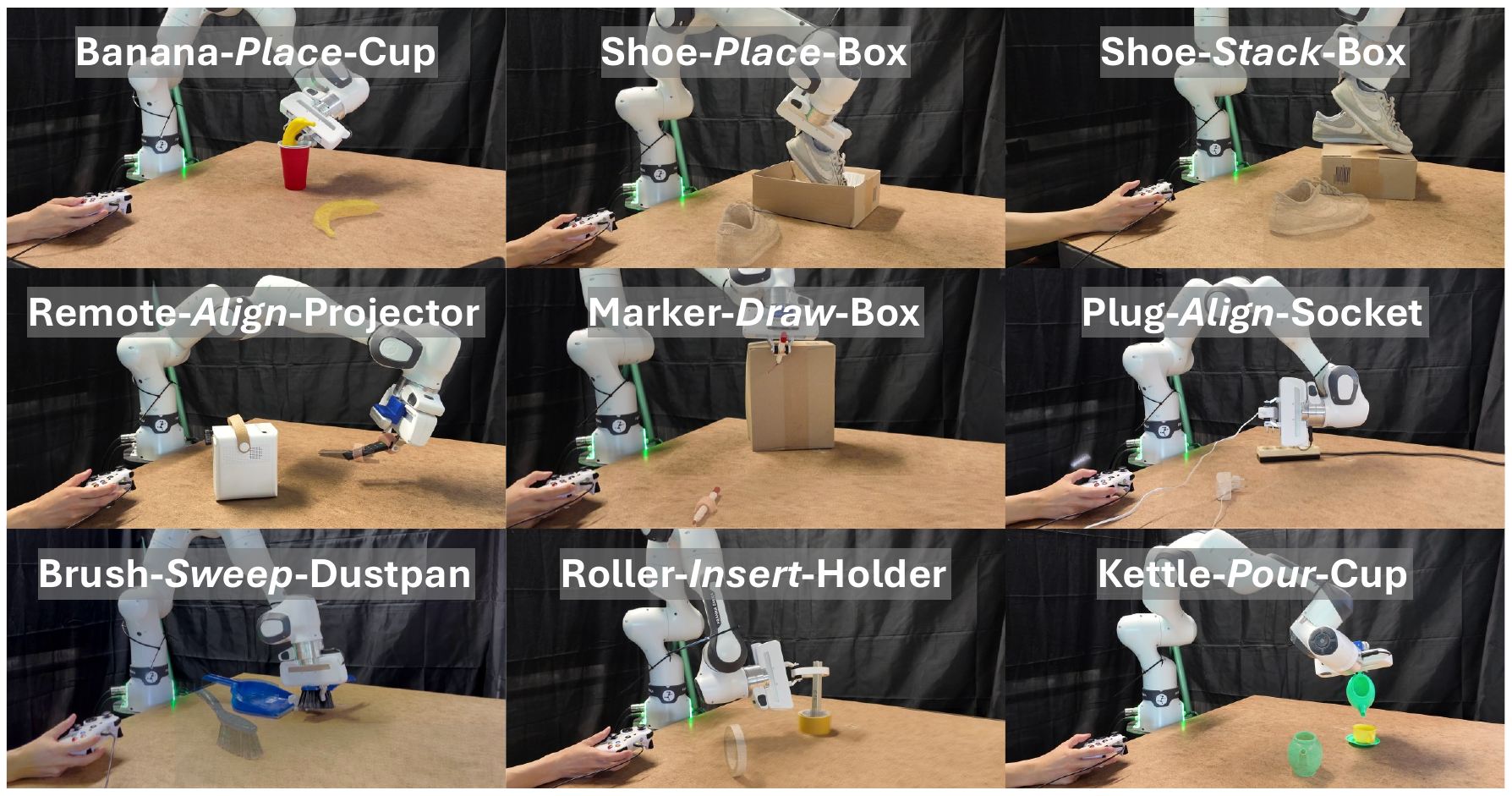}
    \caption{\rmfamily Representative snapshots of successful shared-control task executions with TASC. Semi-transparent objects denote initial positions.}
    \label{fig:real_world_snapshots}
\end{figure}

As illustrated in Fig.~\ref{fig:real_world_snapshots}, TASC demonstrates strong generalization across a wide range of real-world tasks, including placing, aligning, and stacking. Despite significant variations in object geometry and interaction patterns, TASC successfully assists users in completing these tasks without any task-specific tuning. Compared to pure teleoperation, users can complete tasks with one hand by controlling only simple translational motions. 

\begin{figure}[tb]
    \centering
    \includegraphics[width=1.0\linewidth]{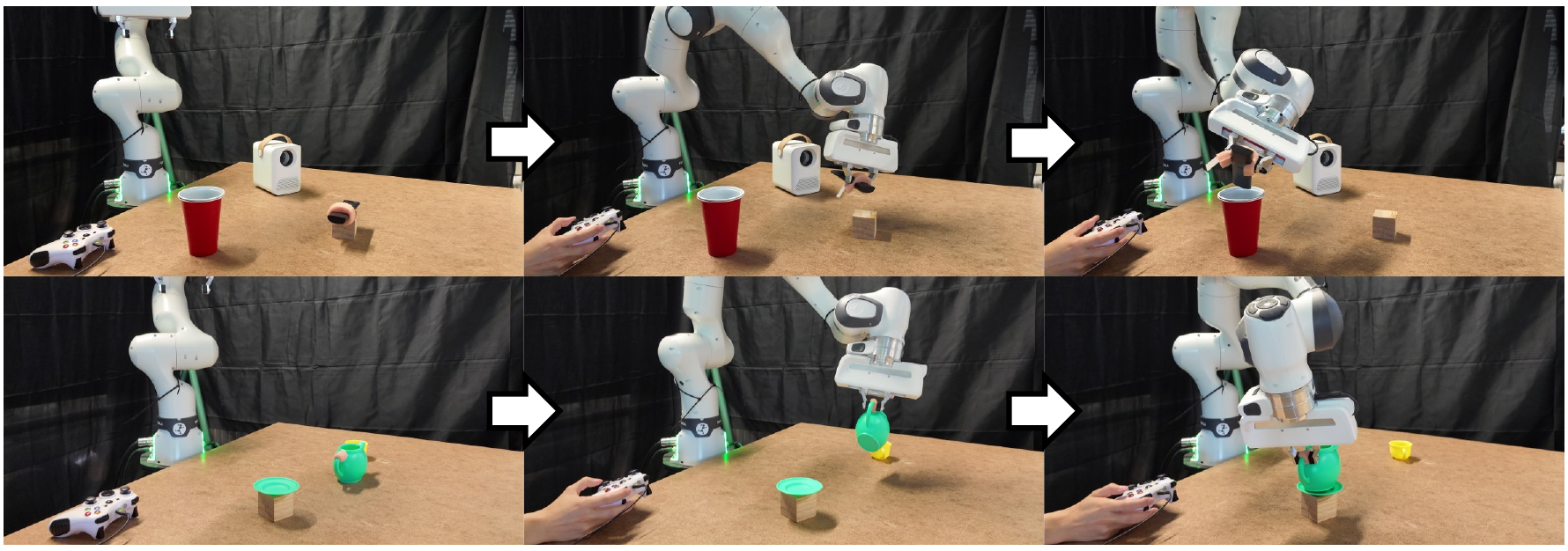}
    \caption{\rmfamily Representative executions of long-horizon multi-step manipulation tasks with TASC. Semi-transparent objects denote initial positions.}
    \label{fig:multi_step}
\end{figure}

We further demonstrate TASC's capability in multi-object and multi-step interactions. As shown in Fig.~\ref{fig:multi_step}, both the remote controller and the kettle interact sequentially with multiple objects. TASC correctly predicts the intended interaction pairs and adapts its assistance accordingly, thereby enabling users to accomplish long-horizon manipulation tasks.

\section{CONCLUSIONS}
We presented TASC, a task-aware shared control framework for relational telemanipulation. Our approach structures task-relevant object relations and spatial affordances into an open-vocabulary interaction graph, enabling assistance during grasping and object interaction. Simulation and real-world experiments demonstrate improved task success rate, reduced completion time, and enhanced user experience compared to existing shared control baselines.
\\
\textbf{Limitation.} First, VLM-based reasoning introduces latency and occasional instability. While future model improvements or local interaction inference~\cite{unmesh2023interacting, huang2024imagination, qi2025two} may alleviate this issue, real-time robustness remains a challenge. Second, our formulation targets prehensile relational manipulation and assumes rigid-object alignment constraints. It focuses on target-pose assistance rather than interaction trajectory generation, and relies on axis-based affordance representations. Extending the framework to richer affordances (e.g., keypoints), non-prehensile actions, and deformable object manipulation remains important future work.


\bibliographystyle{IEEEtran}
\bibliography{ref}

\end{document}